\documentclass[review]{elsarticle}

\usepackage{lineno,hyperref}
\usepackage{amsmath,epsfig}
\usepackage{multirow}
\usepackage{subfigure}
\usepackage{color}
\usepackage{algorithm}
\usepackage{algorithmic}
\usepackage{epstopdf}
\usepackage{appendix}
\usepackage{amsthm}
\usepackage{amssymb}

\modulolinenumbers[5]

\journal{Journal of \LaTeX\ Templates}

\begin{document}

\newcommand{\tabincell}[2]{\begin{tabular}{@{}#1@{}}#2\end{tabular}}

\begin{frontmatter}

\title{Learning Discriminative Features Via Weights-biased Softmax Loss}

\author[Address]{XiaoBin Li}
\ead{lixiaobin161@mails.ucas.ac.cn}

\author[Address]{WeiQiang wang \corref{cor2}}
\ead{wqwang@ucas.ac.cn}

\cortext[cor2]{Corresponding author}
%

\address[Address]{University of Chinese Academy of Sciences \\
Beijing, China}


%



%
%
%
%

\begin{abstract}
Loss functions play a key role in training superior deep neural networks. In convolutional neural networks (CNNs), the popular cross entropy loss together with softmax does not explicitly guarantee minimization of intra-class variance or maximization of inter-class variance. In the early studies, there is no theoretical analysis and experiments explicitly indicating how to choose the number of units in fully connected layer. To help CNNs learn features more fast and discriminative, there are two contributions in this paper. First, we determine the minimum number of units in FC layer by rigorous theoretical analysis and extensive experiment, which reduces CNNs' parameter memory and training time. Second, we propose a negative-focused weights-biased softmax (W-Softmax) loss to help CNNs learn more discriminative features. The proposed W-Softmax loss not only theoretically formulates the intra-class compactness and inter-class separability, but also can avoid overfitting by enlarging decision margins. Moreover, the size of decision margins can be flexibly controlled by adjusting a hyperparameter $\alpha$. Extensive experimental results on several benchmark datasets show the superiority of W-Softmax in image classification tasks.
\end{abstract}


\end{frontmatter}

\section{Introduction}
In the early studies about image classification based on the convolutional neural networks (CNNs), there is no theoretical analysis and experiments explicitly indicating how to choose the number of units in FC layer. If the number of units in FC layer is too big, continued training can result in overfitting of the training data, increasing redundant parameter and training time, and if too small, training can result in underfitting of the training data. FC layer is a form of Artificial Neural Networks (ANN). Early works about determining the number of hidden units for an ANN model\cite{N1994Network,L1998Opt} mainly focus on the size of the training set and the number of input variables, which does not provide theoretical analysis. Without theoretical foundation in the number of nodes of CNNs' FC layer, researchers tend to choose a larger number of nodes. In YOLO\cite{Redmon2015YOLO}, the number of units in FC layer is 1000 for dataset COCO\cite{Lin2014COCO} with 80 classes and RCNN series\cite{R2014Rich,Girshick2015Fast,S2017Faster} network have 4096 units for dataset PASCAL VOC07+12 with 20 classes and dataset COCO, where the number of units is up to 200 times the number of classes. In this paper, we determine the minimum of FC layer number of units by rigorous theoretical analysis and extensive experiments for various classes tasks, which can reduce CNNs' parameter memory and training time.

\begin{figure}[t]
  \centering
  \includegraphics[width=1.0\linewidth]{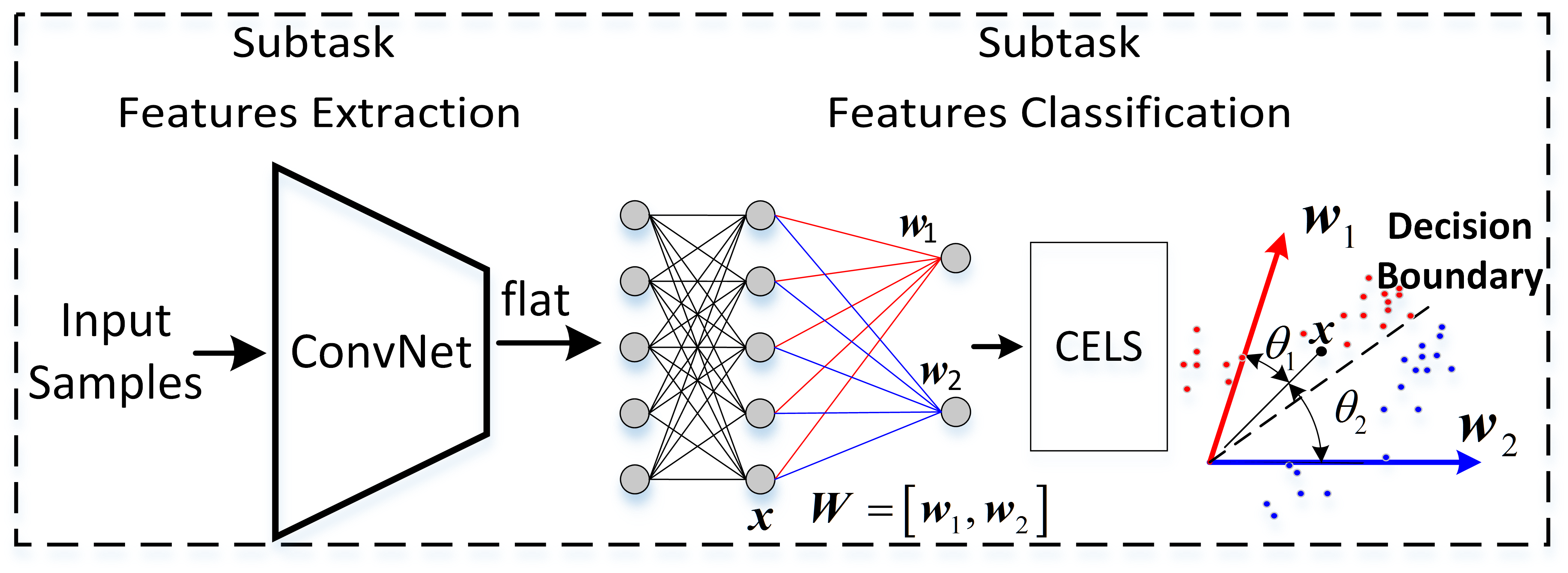}\\
  \caption{Classification task with CNNs. 'CELS' denotes cross entropy loss with softmax.}\label{fig:1}
\end{figure}

In recent years, CNNs have been widely applied in many vision tasks like object recognition and segmentation\cite{Krizhevsky2012ImageNet,Karpathy2014Large,Charles2017PointNet,He2017Mask}, face verification\cite{Taigman2014DeepFace} and hand-writing character recognition\cite{Lecun1998Gradient}. In the CNNs, the convolution layers together with pooling layers are generally used to extract discriminative feature representations, then fully connected layers implement the regression map from features to target labels, i.e., they involve two stages, features extraction and classification, as shown in Fig.\ref{fig:1}.

In the aspect of feature representation learning, many effective techniques have been presented during the past decade. For example, the deeper and wider network architectures are built to improve the performances of CNNs\cite{He2016Deep}\cite{Szegedy2014Going}; different feature normalizations are adopted, like batch normalization\cite{Ioffe2015Batch}, layer normalization\cite{Ba2016Layer}, instance normalization\cite{Ulyanov2016Instance} and group normalization\cite{Wu2018Group}; diverse non-linear activation functions are exploited, like PReLU\cite{He2015Delving}; weights regularization~\cite{Wan2013Regularization} and stochastic pooling~\cite{Matthew2013Stochastic} are also investigated. However, all these techniques play a supporting role in extracting features fast and accurately, since the training of network is driven by loss calculated in fully connected (FC) layer. Now, overfitting is still a challenge to be addressed for CNNs.

In features classification subtask, FC layer with softmax loss is the mainstream where softmax loss tends to makes CNNs early stop in training. Actually, softmax function is sensitive to the size of input values, which is the main weakness of softmax loss. For example, considering the binary classification(referring to Fig.\ref{fig:weights} $C=2$ case), the decision boundary of conventional Softmax loss is depicted as $y_1=y_2(y_1=\|\emph{\textbf{w}}_1\|\|\emph{\textbf{x}}\|\cos{(\theta_1)}, y_2=\|\emph{\textbf{w}}_2\|\|\emph{\textbf{x}}\|\cos{(\theta_2)})$, where $\emph{\textbf{w}}_1$ and $\emph{\textbf{w}}_2$ denote the weight vectors of two classes, $\emph{\textbf{x}}$ denotes the feature representation for a given instance, $\theta_1$ and $\theta_2$ are the angles between weight vectors and feature. Here we suppose $\|\emph{\textbf{w}}_1\|=\|\emph{\textbf{w}}_2\|=1$, $\cos{(\theta_1)}=0.05$ and $\cos{(\theta_2)}=-0.05$, which means feature $\emph{\textbf{x}}$ is very close to the decision boundary. When we increase the norm of feature $\emph{\textbf{x}}$, e.g., let $\|\emph{\textbf{x}}\|$ equal 1, 10, 30, 50 respectively, $softmax([y_1, y_2])$= [0.52, 0.48], [0.73, 0.27], [0.95, 0.05], [0.99, 0.01] correspondingly. So, feature $\emph{\textbf{x}}$ with a large norm makes the Softmax loss decrease easily to zero, even if $\theta_1$ is approximately equals to $\theta_2$. The distribution of features makes CNNs train easily and perform poorly in testing and that is where the inertia of CNNs exists.

The weakness of CNNs that softmax loss does not rigorously encourage intra-class compactness and inter-class separability is revealed by experiments. To overcome this  problem, many research works have been carried out\cite{Wen2016A,Chen2014Deep,F2015FaceNet,Yang2018Robust,Liu2016Large,Liu2017SphereFace}. All these studies focus on encouraging better discriminating performance: minimizing intra-class variance and maximizing inter-class variance. Wen \emph{et al.}\cite{Wen2016A} proposed the center loss and used Euclidean distance to measure the distance between two instances, in which the input must be a pair of instances. It does not explicitly encourage the inter-class separability, which still not gets rid of overfitting. Chen \emph{et al.}\cite{Chen2014Deep} proposed contrastive loss and set hyper-parameter $margin$ to train Siamese network, in which the input pairs should be careful selected ones from training sets. Similar to cite{Chen2014Deep}, Schroff \emph{et al.}\cite{F2015FaceNet} proposed triplet loss to learn more discriminating representation in which the input triplets need to be designed too. Yang \emph{et al.}\cite{Yang2018Robust} proposed prototype learning to increase CNNs robustness, however, prototype learning totally abandoned softmax layer. \cite{Liu2016Large} and \cite{Liu2017SphereFace}  proposed large-margin softmax (L-Softmax) loss and angular softmax (A-Softmax) loss respectively, which transfer Euclidean margin learning to angular margin learning. While both L-Softmax and A-Softmax can be optimized by typical stochastic gradient descent, they design complicated function $\psi{(\theta)}$ based on angle, resulting in increased difficulty and time in training. The training difficulty and hyper-parameter $m$ in L-Softmax and A-Softmax are positive correlation.
\begin{figure}[t]
\centering
\subfigure[original softmax]{
\label{fig:orig}
\begin{minipage}{0.49\linewidth}
\centering
\includegraphics[width=1.0\linewidth]{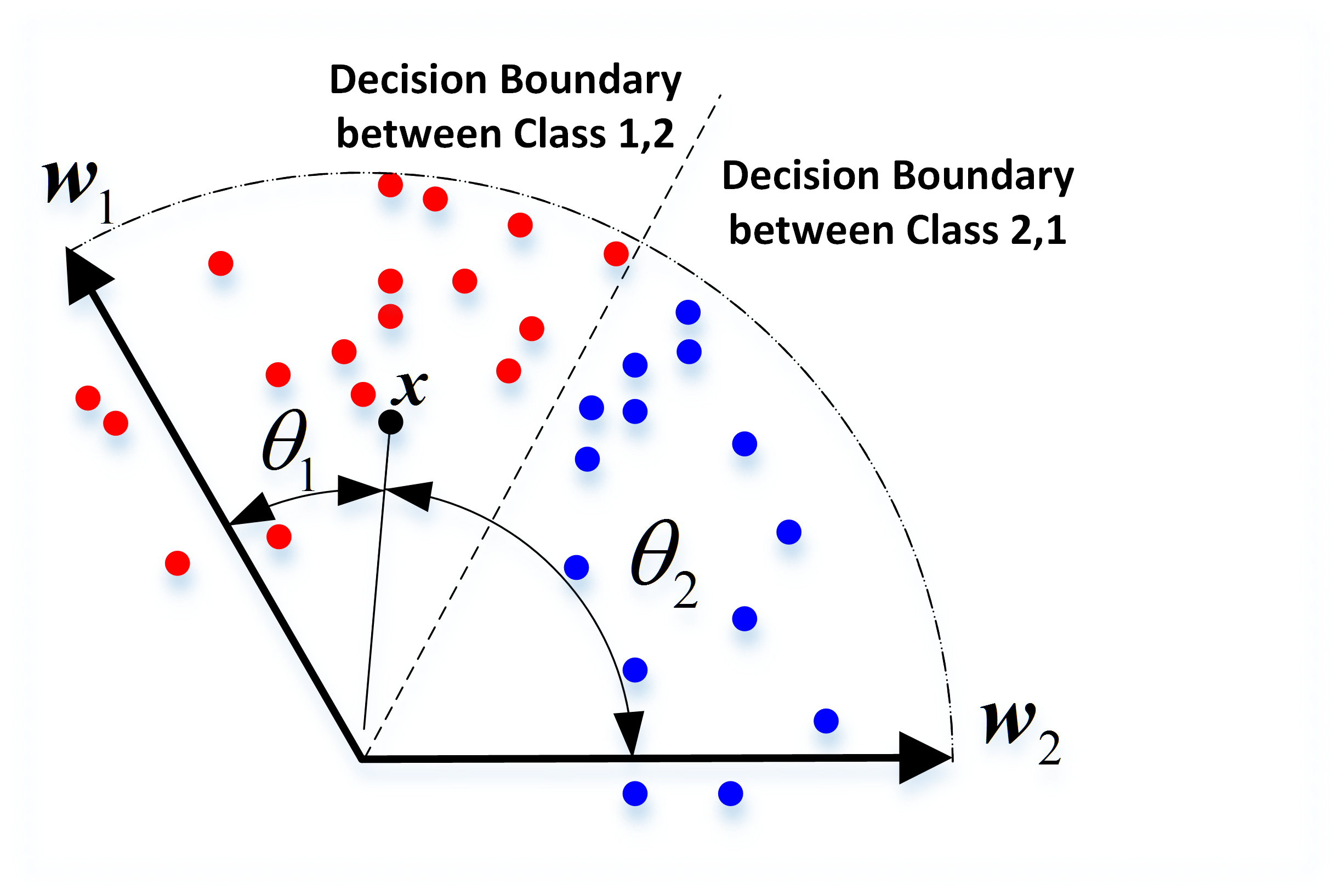}
\end{minipage}%
}%
\subfigure[$\alpha=0.5$]{
\label{fig:05}
\begin{minipage}{0.49\linewidth}
\centering
\includegraphics[width=1.0\linewidth]{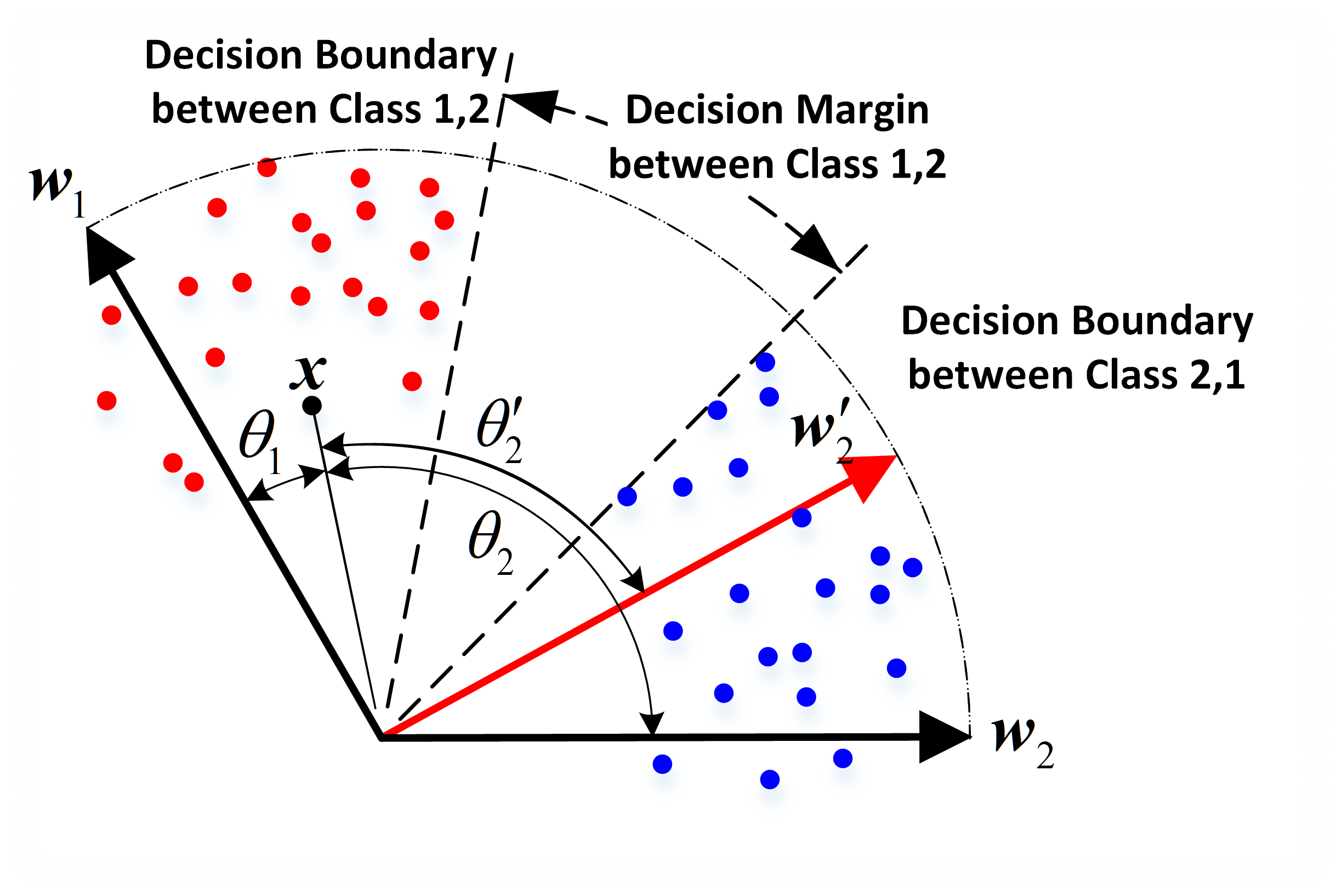}
\end{minipage}%
}%

\subfigure[$\alpha=1.0$]{
\label{fig:10}
\begin{minipage}{0.49\linewidth}
\centering
\includegraphics[width=1.0\linewidth]{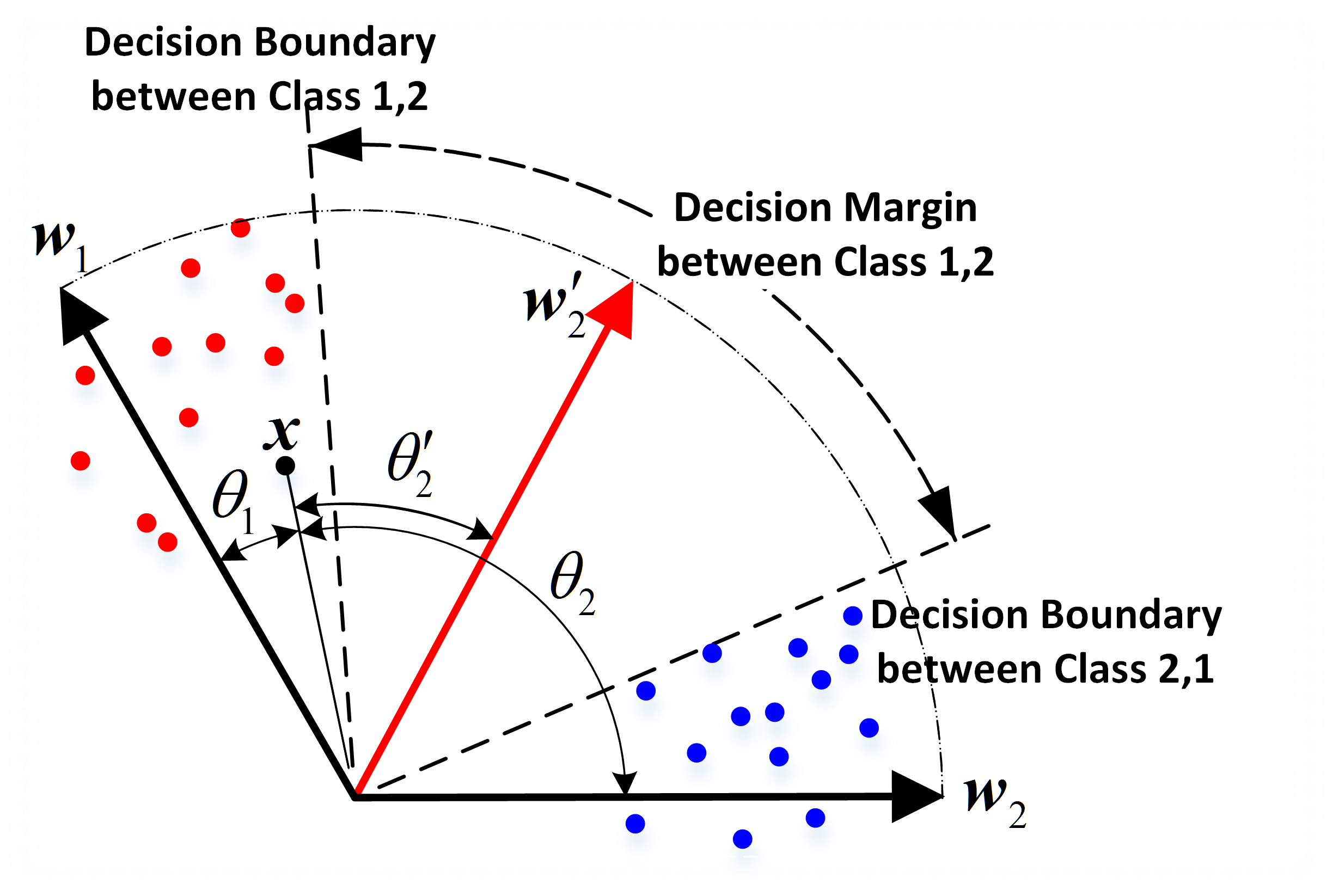}
\end{minipage}
}%
\subfigure[$\alpha=1.5$]{
\label{fig:15}
\begin{minipage}{0.49\linewidth}
\centering
\includegraphics[width=1.0\linewidth]{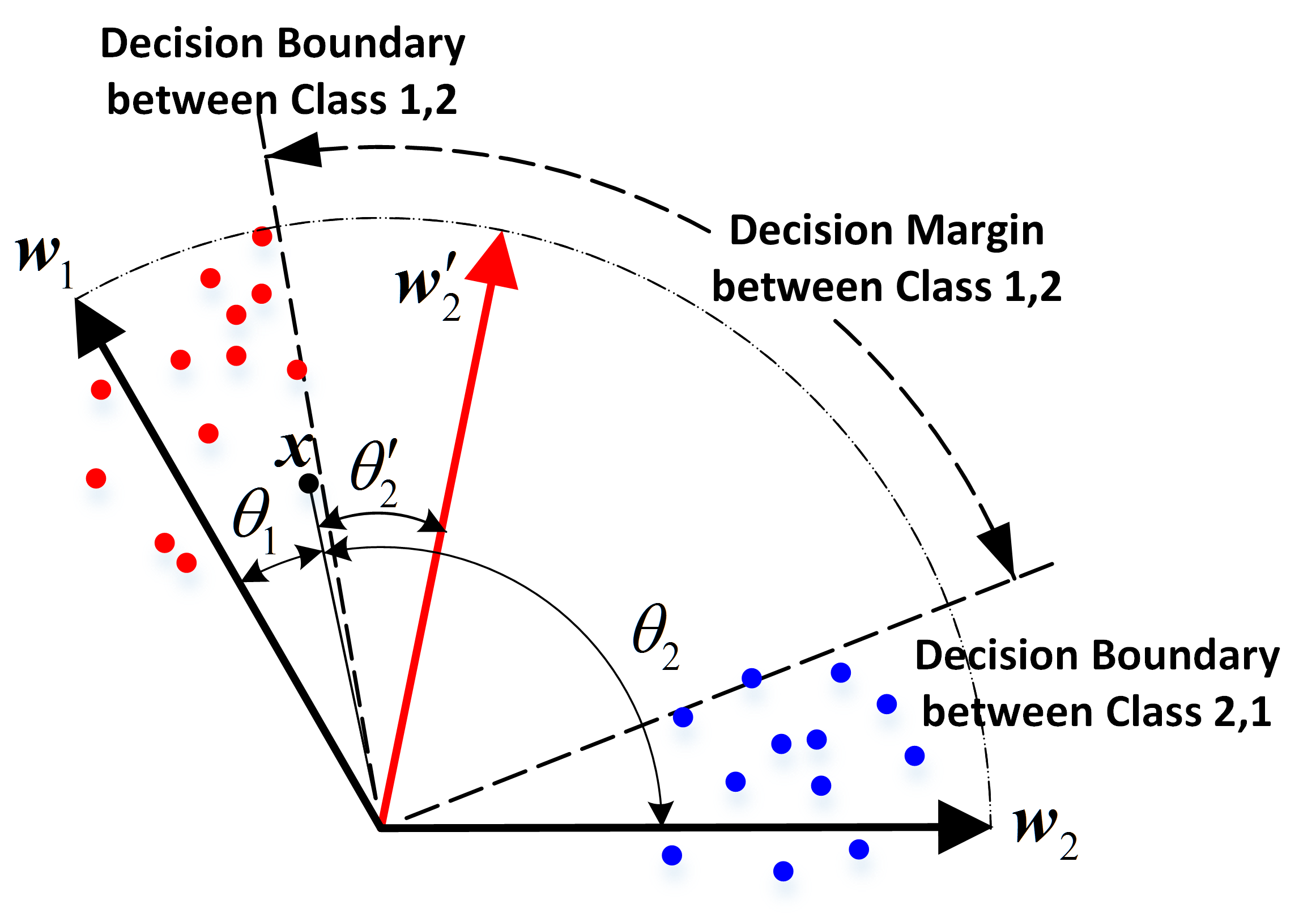}
\end{minipage}
}%
\centering
\caption{The comparison of original softmax loss and W-Softmax loss when training instances with label 1. (a) is original softmax loss, where the decision boundary is coincident, and (b)-(d) are W-Softmax loss, where the decision boundaries get separated and $\emph{\textbf{w}}'_2=\frac{\alpha\emph{\textbf{w}}_1+\emph{\textbf{w}}_2}{\|\alpha\emph{\textbf{w}}_1+\emph{\textbf{w}}_2\|}$.}
\label{fig:3}
\end{figure}

We propose a new Softmax-like loss function, called the negative-focused weights-biased softmax (W-Softmax) loss, which has no extra trainable parameters compared with the conventional Softmax loss. By increasing the probabilities of all the negative classes in the softmax output, W-Softmax loss can help CNNs learn more discriminative features. Generally, while training $c$-th class instances in the multi-class classification, each $\emph{\textbf{w}}_i(i\neq{c})$ is replaced by normalized $\alpha\emph{\textbf{w}}_c+\emph{\textbf{w}}_i$ so that the decision boundaries between any two classes get separated and the decision margins are enlarged. Fig.\ref{fig:3} illustrates the idea via the example of two-class classification, where the decision margin increases when hyper-parameter $\alpha$ gets bigger. When training CNNs using the conventional softmax loss, the decision boundary between any two classes is coincident, and it brings premature convergence of CNNs in training when features distribute around the decision boundaries. However, by using the proposed the W-Softmax loss, the problem can be addressed, since the loss of CNNs will be enlarged if features locate around the original decision boundaries, separating the decision boundaries and enlarging the decision margins.

The W-Softmax loss can force features to draw close to the weight vectors of their corresponding class by increasing the value of $\alpha (\alpha\geq{0})$. A bigger $\alpha$ corresponds to a larger decision boundary margin, and the strong constraint tends to make intra-class variance decrease and inter-class variance increase. Compared with other works \cite{Liu2016Large,Liu2017SphereFace}, the proposed loss function does not need to calculate the cosine values and use multiple-angle formula, thus it is computationally very efficient in the training and optimization, just as the conventional Softmax loss. In fact, the softmax loss is a special case of W-Softmax loss when $\alpha=0$. The contributions of this work are summarized as follows:
\begin{enumerate}
  \item We determine the minimum number of units in FC layer by rigorous theoretical analysis and extensive experiments for various classes tasks, which reduces CNNs' parameter memory and training time.
  \item We present a new W-Softmax loss to make CNNs learn more discriminative features, and it can effectively improve the classification performance by avoiding premature convergence.
  \item The size of decision margins can be optionally adjusted by a positive real-value paremeter $\alpha$. By increasing the value of $\alpha$, CNNs can maximize inter-class variance and minimize intra-class variance. Extensive experiments on benchmark datasets show the effectiveness of W-Softmax loss .
\end{enumerate}


\section{Related Works}

\subsection{Units in Artificial Neural Networks}
In image classification task based on convolutional neural networks (CNNs), fully connected (FC) layer is a common method in feature classification. In the early studies without CNNs, artificial neural networks (ANNs) are the focus in artificial intelligence and pattern recognition, where FC layer is a form of ANNs. Murata \emph{et al.}\cite{N1994Network} studied the relation between the training error and the generalization error in terms of the number of the training examples and the complexity of a network which reduces to the number of parameters in the ordinary statistical theory of Akaike's information criterion (AIC). The number of hidden units is selected based on a given training set. Fletcher \emph{et al.}\cite{L1998Opt} developed an algorithm to optimize the number of hidden nodes in  feedforward artificial neural network by minimizing the mean square error over noisy training data, where network minimized the number of training sessions necessary for optimization of the number of hidden nodes. All these works focused on optimizing the number of hidden nodes of the whole ANNs. Differently, this paper is aimed at determining the number of units in FC layer for image classification task based on CNNs. The minimum number of FC layer units is determined by rigorous theoretical analysis and extensive experiments, which shows the minimum number varies from various classes in image classification.
\subsection{Loss Functions}
The design of loss functions plays a significant role in training deep networks. Various loss functions have been presented and applied to learn discriminating feature representations. Contrastive loss \cite{Chen2014Deep} and triplet loss\cite{F2015FaceNet} need to carefully select instance pairs and triplet instances as the input of network in the train stage, since the performance of CNNs heavily depends on selected training instances. Similar to contrastive loss and triplet loss in increasing the Euclidean margin, Yang \emph{et al.}\cite{Yang2018Robust} is a kind of k-nearest-neighbor (K-NN) method, which totally abandons the softmax layer and increases the burden of storages space and computation requirement. Center loss\cite{Wen2016A} together with softmax loss can help CNNs reduce the intra-class variance and learn more discriminative features. Liu \emph{et al.}\cite{Liu2016Large} proposed a large-margin softmax loss and designed an angle function $\psi{(\theta)}$ related to $m$ to decrease the probability of positive instances, which improves the feature discrimination. Liu \emph{et al.}\cite{Liu2017SphereFace} used A-Softmax loss to encourage a large angular margin similar to L-Softmax. Differently, the A-Softmax loss normalized the weights by $L_2$-norm, which has demonstrated its effectiveness on a series of open-set face recognition benchmarks. Both L-Softmax loss and A-Softmax loss are positive-focused softmax loss since both of them decrease the probability of positive class by enlarging the angle between features and weight vectors of positive class. However, when the integer hyper-parameter $m$ ($m=2,3,4...$) is too big, the training of CNNs become very difficult.

Compared with the L-Softmax and A-Softmax losses, the proposed W-Softmax loss is a negative-focused softmax loss. We first remove the biases from the last FC layer and normalize weight vectors of all the classes by $L_2$-norm and then evaluate the weight vectors of each negative class by Eq.(\ref{eq:3}).
\begin{equation}\label{eq:3}
    \emph{\textbf{w}}'_i=\frac{\alpha\emph{\textbf{w}}_c+\emph{\textbf{w}}_i}{\|\alpha\emph{\textbf{w}}_c+\emph{\textbf{w}}_i\|}, (i\neq{c})
\end{equation}
where $c$ is the index of positive classifier weight vector, $i$ is the index of negative classifier weight vector, $\emph{\textbf{w}}_c$ is positive classifier weight vector and $\emph{\textbf{w}}_i$ is negative classifier weight vector. When training instance with label $c$, the weight matrix in the last FC layer is transformed as $\emph{\textbf{W}}'=[\emph{\textbf{w}}'_1,\cdots,\emph{\textbf{w}}'_{c-1},\emph{\textbf{w}}_c,\emph{\textbf{w}}'_{c+1},\cdots\emph{\textbf{w}}'_{C}]$, where only the positive weight vector with true label $c$ is not transformed. After the inner product between $\emph{\textbf{W}}'$ and $\emph{\textbf{x}}$, we get the output of the last FC layer $\emph{\textbf{f}}=\emph{\textbf{W}}'^T\emph{\textbf{x}}$. And then $\emph{\textbf{f}}$ is input into the softmax layer and the softmax loss is calculated the same as original softmax loss. In testing time, we use original classifier weight matrix $\emph{\textbf{W}}=[\emph{\textbf{w}}_1,\cdots,\emph{\textbf{w}}_{c-1},\emph{\textbf{w}}_c,\emph{\textbf{w}}_{c+1},\cdots \emph{\textbf{w}}_{C}]$ instead. In this case, we encourage the negative classes probabilities in softmax and increase their loss, which makes CNNs more stricter with the positive class and learn more discriminating features.

\section{Determining the Number of Units in FC Layer}
\begin{figure}[t]
\begin{center}
  \includegraphics{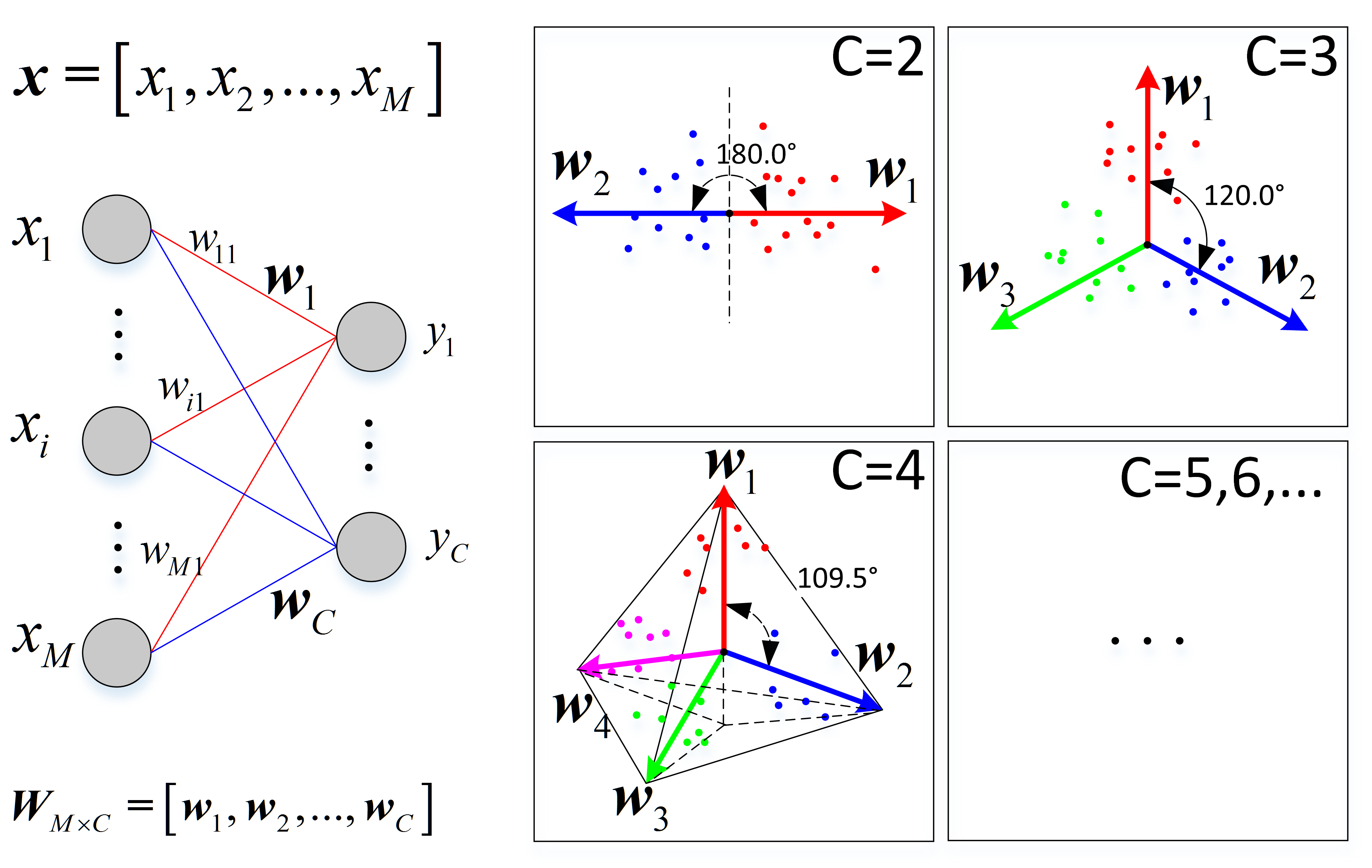}
\end{center}
  \caption{The optimal weight vectors' distribution in classification tasks with various classes, where the angle between any two weight vectors is equal. $C$ is the number of classes and $M$ is the length of feature vector $\emph{\textbf{x}}$. $W_{M\times{C}}$ is classifier weight matrix and column vector $\emph{\textbf{w}}_i$ is classifer weight vector for class $i$. $y_i(y_i=\emph{\textbf{w}}_i^T\emph{\textbf{x}})$ is the output of class $i$.}
  \label{fig:weights}
\end{figure}

For $C$-classes classification task, the feature vector extracted from convolutional network is $\emph{\textbf{x}}$ with length $M$ and classifier weight matrix without biased in FC layer is $\emph{\textbf{W}}_{M\times{C}}=[\emph{\textbf{w}}_1,\emph{\textbf{w}}_2,...,\emph{\textbf{w}}_C]$ denoted as $\emph{\textbf{W}}_C$ seen in Fig.\ref{fig:weights}. Geometrically, vector $\emph{\textbf{x}}$ and $\emph{\textbf{w}}_i$ are the points in $\mathbb{R}^M$ (i.e. $\emph{\textbf{x}}, \emph{\textbf{w}}_i \in \mathbb{R}^M$). Theoretically, the distribution of weight vectors is optimal when the weight vectors are uniformly distributed in space, which means the angle between any two weight vectors is a constant value. To facilitate analysis, all the weight vectors are normalized by $L_2$. For $i=1,2,\cdots,C$, $\emph{\textbf{w}}_i=[w_{1i},w_{2i},...,w_{Mi}]^T$ and $\|\emph{\textbf{w}}_i\|=1$. In the feature vector space $\mathbb{R^M}$, all the weight vectors are points on the hyper unit sphere and for all index $i, j$ and $k$ ($i\neq{j}, i\neq{k}, j\neq{k}$), there exists $d(\emph{\textbf{w}}_i,\emph{\textbf{w}}_j)={d(\emph{\textbf{w}}_i,\emph{\textbf{w}}_k)}$, where $d(*)$ is Euclidean distance function. So the solution is turned to how to determine the range of variable $M$ to ensure that the problem that the angle between any two weight vectors is a constant value has a solution.

 The problem is formulated by
\begin{align}
 \emph{\textbf{w}}_2^T\emph{\textbf{w}}_1=\emph{\textbf{w}}_3^T\emph{\textbf{w}}_1=...=\emph{\textbf{w}}_C^T\emph{\textbf{w}}_1 \nonumber \\
 \label{eq:weightsequal}=\emph{\textbf{w}}_3^T\emph{\textbf{w}}_2=...=\emph{\textbf{w}}_C^T\emph{\textbf{w}}_2  \\
 \vdots \nonumber \\
 =\emph{\textbf{w}}_{C-1}^T\emph{\textbf{w}}_C, \nonumber
\end{align}
The minimum value of $M$ denoted as $M_{min}$ for $C$-classes task can be determined by mathematical induction. For continuously increasing number $C$, the weight matrix $\emph{\textbf{W}}_C$ is constructed from a special solution $\emph{\textbf{W}}_2$ of Eq.\ref{eq:weightsequal}, which is
\begin{equation}\label{eq:MI}
\emph{\textbf{W}}_C=\left\{
\begin{array}{ccc}
\left[\begin{array}{cc}
1&-1
\end{array}\right]_{1\times{2}} & &  C=2 \\
 & &  \\
\left[\begin{array}{cc}
\sqrt{\frac{(C-1)^2-1}{(C-1)^2}}\emph{\textbf{W}}_{C-1}&\emph{\textbf{0}}\\
-\frac{\emph{\textbf{1}}}{\emph{\textbf{C-1}}}&1
\end{array}\right]_{(C-1)\times{C}} & & C>2
\end{array}\right.
\end{equation}
where each weight matrix $\emph{\textbf{W}}_C$ satisfies the condition in Eq.\ref{eq:weightsequal} and for all $i$, $\|\emph{\textbf{w}}_i\|=1$, and there exists
 \begin{equation}\label{eq:weightij}
\emph{\textbf{w}}_i^T\emph{\textbf{w}}_j=
\begin{cases}
1& i=j,\\
-\frac{1}{C-1} & i\neq{j}.
\end{cases}
\end{equation}
 In Eq.\ref{eq:MI}, the distribution of $\emph{\textbf{W}}_2$, $\emph{\textbf{W}}_3$ and $\emph{\textbf{W}}_4$ are the cases $C=2,3,4$ respectively shown in Fig.\ref{fig:weights}. Because the special case $\emph{\textbf{W}}_2(M=1)$ in Eq.\ref{eq:MI} is the simplest case and the size of weight matrix $\emph{\textbf{W}}_C$ is $M\times{C}$, we can determine the minimum value of $M$ as $M_{min}=C-1$, which means if $M\geq{C-1}$, the Eq.\ref{eq:weightsequal} has solution.

 Next, we will prove that the $C-1$ is the minimum value for $M$ in construction. Reductio ad absurdum is adopted to prove the assumption.
\newtheorem{assumption}{Assumption}
\begin{assumption}\label{as:ap1}
There is no unit vector $\emph{\textbf{w}}_{C+1}(C>2)$ with length $C-1$ making new weight matrix $\emph{\textbf{W'}}_{C+1}=[\emph{\textbf{W}}_C,\emph{\textbf{w}}_{C+1}]$ satisfy Eq.\ref{eq:weightsequal}, where $\emph{\textbf{W}}_C$ is from Eq.\ref{eq:MI}.
\end{assumption}

\begin{proof}[\textbf{Proof}]
Suppose there is a unit vector $\emph{\textbf{w}}_{C+1}$ with length $C-1$ making new weight matrix $\emph{\textbf{W'}}_{C+1}=[\emph{\textbf{W}}_C,\emph{\textbf{w}}_{C+1}]=[\emph{\textbf{w}}_{1},\cdots,\emph{\textbf{w}}_{C},\emph{\textbf{w}}_{C+1}]$ satisfy Eq.\ref{eq:weightsequal}, described as
\begin{equation}\label{eq:proof1}
 \emph{\textbf{w}}_1^T\emph{\textbf{w}}_{C+1}=\emph{\textbf{w}}_2^T\emph{\textbf{w}}_{C+1}=\dots=\emph{\textbf{w}}_{C-1}^T\emph{\textbf{w}}_{C+1}=\emph{\textbf{w}}_C^T\emph{\textbf{w}}_{C+1}.
\end{equation}
According the construction, the rank of weight matrix $\emph{\textbf{W}}_C$ is $C-1$, there exits nonzero vector $\emph{\textbf{a}}=[a_1,a_2,\cdots,a_{C-1}]$ satisfying $\emph{\textbf{w}}_C=a_1\emph{\textbf{w}}_1+a_2\emph{\textbf{w}}_2+\cdots+a_{C-1}\emph{\textbf{w}}_{C-1}$. The same as $\emph{\textbf{w}}_C$, there exits nonzero vector $\emph{\textbf{b}}=[b_1,b_2,\cdots,b_{C-1}]$ satisfying
\begin{equation}\label{eq:proof2}
\emph{\textbf{w}}_{C+1}=b_1\emph{\textbf{w}}_1+b_2\emph{\textbf{w}}_2+\cdots+b_{C-1}\emph{\textbf{w}}_{C-1}.
\end{equation}
Substituting Eq.\ref{eq:proof2} into Eq.\ref{eq:proof1} and then simplifying the equation by Eq.\ref{eq:weightij}, we gets
\begin{align}
b_1-\frac{1}{C-1}(b_2+b_3+\cdots+b_{C-1})&=b_2-\frac{1}{C-1}(b_1+b_3+\cdots+b_{C-1})\nonumber\\
\label{eq:proof3}&\vdots\\
&=b_{C-1}-\frac{1}{C-1}(b_1+b_2+\cdots+b_{C-2})\nonumber\\
&=-\frac{1}{C-1}(b_1+b_2+\cdots+b_{C-1}).\nonumber
\end{align}
Because $C>2$, Eq.\ref{eq:proof3} has only one solution $b_1=b_2=\cdots=b_{C-1}=0$, which is inconsistent with the assumption. Therefore, according to Reductio ad absurdum, assumption\ref{as:ap1} is correct, which means that the length of weight vector $\emph{\textbf{w}}_{C+1}$ is at least $C$. Under the condition of \ref{eq:weightsequal}, the construction in Eq.\ref{eq:MI} ensures that each weight vector $\emph{\textbf{w}}_{i}$ in weight matrix $\emph{\textbf{W}}_{C}$ has the minimum length($C-1$), in other words $M_{min}=C-1$.
\end{proof}

Extensive experiments on many benchmark datasets validate our conclusion.

\section{Weights-biased Softmax Loss}
\subsection{Review of Conventional Softmax Loss}
In this section, we review the conventional softmax loss. Suppose we have a $C$-classes classification task. For a given instance with label $c$, its feature is $\emph{\textbf{x}}$. The probability for every class can be evaluated by
\begin{equation}\label{eq:4}
\begin{split}
  p_i=\frac{\exp{(\emph{\textbf{w}}_i^T\emph{\textbf{x}}+b_i)}}{\sum_{j=1}^{C}\exp{(\emph{\textbf{w}}_j^T\emph{\textbf{x}}+b_j)}},
\end{split}
\end{equation}
where $\emph{\textbf{w}}_i$ and $b_i$ denote the weights and biases of the last FC layer.
In the prediction stage, an instance is classified to label $i$ if $p_i>p_j$ (for all $j$ and $j\neq{i}$). It can be converted as  $\emph{\textbf{w}}_i^T\emph{\textbf{x}}+b_i>\emph{\textbf{w}}_j^T\emph{\textbf{x}}+b_j$, i.e.,  $\|\emph{\textbf{w}}_i\|\|\emph{\textbf{x}}\|\cos{\theta_i}+b_i>\|\emph{\textbf{w}}_j\|\|\emph{\textbf{x}}\|\cos{\theta_j}+b_j$, where $\theta_i$ denotes the angle between $\emph{\textbf{w}}_i$ and $\emph{\textbf{x}}$, $0\leq{\theta_i}\leq{\pi}$. The decision boundary of two classes $i$ and $j$ is defined by $\|\emph{\textbf{w}}_i\|\|\emph{\textbf{x}}\|\cos{\theta_i}+b_i=\|\emph{\textbf{w}}_j\|\|\emph{\textbf{x}}\|\cos{\theta_j}+b_j$. If we let $\|\emph{\textbf{w}}_i\|=1$ and remove the biases, the decision boundaries become $\cos{\theta_i}=\cos{\theta_j}$, so the angle between weight vector $\emph{\textbf{w}}_i$ of each class and feature $\emph{\textbf{x}}$ is very important for classification.

The multi-class softmax loss for an instance $\emph{\textbf{x}}$ can be formulated by
\begin{equation}\label{eq:6}
\begin{aligned}
  L&=-\log{p_c}=-\log{(\frac{\exp{(\emph{\textbf{w}}_c^T\emph{\textbf{x}}+b_c)}}{\sum\limits_{j=1}^C{\exp{(\emph{\textbf{w}}_j^T\emph{\textbf{x}}+b_j)}}})} \\
  &=-\log{(\frac{\exp{(\|\emph{\textbf{w}}_c\|\|\emph{\textbf{x}}\|\cos{\theta_c}+b_c)}}{\sum\limits_{j=1}^C{\exp{(\|\emph{\textbf{w}}_j\|\|\emph{\textbf{x}}\|\cos{\theta_j}+b_j)}}})},
\end{aligned}
\end{equation}
where $c$ is the class label of instance $\emph{\textbf{x}}$. The decision boundary for class $c$ and class $i$ can be defined by  $\|\emph{\textbf{w}}_c\|\|\emph{\textbf{x}}\|\cos{\theta_c}+b_c=\|\emph{\textbf{w}}_i\|\|\emph{\textbf{x}}\|\cos{\theta_i}+b_i (i\neq{c})$. Because the decision boundary between two classes is coincident, the conventional softmax loss cannot make CNNs learn a more discriminative feature representation. To encourage the ability of feature representation, we propose a new weights-biased softmax loss.

\subsection{Weights-Biased Softmax Loss}
\textbf{Positive Probability and Negative Probabilities}. To obtain a large decision margin, we present a new loss called the weights-biased Softmax loss (W-Softmax), which utilizes parameter $\alpha$ in Eq.(\ref{eq:3}) to control the size of expected decision margin. Fig.\ref{fig:3} illustrates the basic principle of the proposed loss via an example of two-class classification, and it is also true for the case of  multi-class classification. In our CNN network, we first remove the biases in the last FC layer of CNN and normalize the corresponding weight vectors, i.e., let $\|\emph{\textbf{w}}_i\|=1$. For a given class $c$ corresponding to $\emph{\textbf{w}}_c$, other classes are called negative class and each negative class $i (i\neq c)$ has a corresponding weight $\emph{\textbf{w}}'_i$ evaluated by Eq.(\ref{eq:3}), which is specifically used for evaluating the loss of instance $\emph{\textbf{x}}$ in class $c$. It should be noted that, for two vectors $\emph{\textbf{p}},\emph{\textbf{q}}$ with angle $\theta\in[0,\pi]$ between them, if $\alpha>0$ and $\emph{\textbf{z}}=\alpha\emph{\textbf{p}}+\emph{\textbf{q}}$, then vector must fall into angle $\theta$, and vector $\emph{\textbf{z}}$ will get closer to vector $\emph{\textbf{p}}$ as $\alpha$ gets larger. So, $\emph{\textbf{w}}'_i$ must fall into the included angle of $\emph{\textbf{w}}_c$ and $\emph{\textbf{w}}_i$. Further, the angular bisector of $\emph{\textbf{w}}_c$ and $\emph{\textbf{w}}'_i$ forms a new decision boundary, which makes instances from class $c$ become closer to $\emph{\textbf{w}}_c$ and far away from $\emph{\textbf{w}}_i$ in training. In $C$-classes classification, for input feature $\emph{\textbf{x}}$ to the last FC layer with label $c$, the positive probability $p_c$ and negative probabilities $p_i(i\neq{c})$ are evaluated by
\begin{equation}\label{eq:7}
\begin{split}
  p_c=\frac{\exp{(\|\emph{\textbf{x}}\|\cos{\theta_c})}}{\exp{(\|\emph{\textbf{x}}\|\cos{\theta_c})}+\sum\limits_{j\neq{c}}^{C}\exp{(\|\emph{\textbf{x}}\|\cos{\theta'_j})}}, \\
  p_i=\frac{\exp{(\|\emph{\textbf{x}}\|\cos{\theta'_i})}}{\exp{(\|\emph{\textbf{x}}\|\cos{\theta_c})}+\sum\limits_{j\neq{c}}^{C}\exp{(\|\emph{\textbf{x}}\|\cos{\theta'_j})}},
\end{split}
\end{equation}
where $\theta_c$ and $\theta'_i$ denote the angles between weight vector $\emph{\textbf{w}}_c$ and feature $\emph{\textbf{x}}$, as well as $\emph{\textbf{w}}'_i$ and $\emph{\textbf{x}}$, respectively.

\textbf{Decision Boundaries for Class $c$}. In the training phase of learning features, for instance in class $c$, we use $\emph{\textbf{W}}'=[\emph{\textbf{w}}'_1,\cdots, \emph{\textbf{w}}'_{c-1},\emph{\textbf{w}}_c,\emph{\textbf{w}}'_{c+1},\cdots \emph{\textbf{w}}'_{C}]$ to characterize the boundaries between class $c$ and other classes. Concretely, let $p_c=p_i(i\neq{c})$, we can easily derive $\theta_c=\theta'_i$, which means the decision boundary between class $c$ and class $i$ is the angular bisector of angle between $\emph{\textbf{w}}_c$ and $\emph{\textbf{w}}'_{i}$. In testing phase, $\emph{\textbf{W}}'$ is replaced by original weights $\emph{\textbf{W}}=[\emph{\textbf{w}}_1, \emph{\textbf{w}}_2,...,\emph{\textbf{w}}_c,..., \emph{\textbf{w}}_{C-1}, \emph{\textbf{w}}_{C}]$. Fig.\ref{fig:3} illustrates the decision boundaries for two classes, where the class 1 is considered as the positive class.
In the conventional Softmax loss, the decision boundary margin is zero, so learned features from two classes likely distribute very close on both sides of their common decision boundary. In W-Softmax loss, the decision margins are magnified by parameter $\alpha$ in Eq.\ref{eq:3}. It is clear that there exists $\theta_2=\theta'_2+\theta_{\emph{\textbf{w}}'_2, \emph{\textbf{w}}_2}$, where $\theta_{\emph{\textbf{w}}'_2, \emph{\textbf{w}}_2}$ is the angle between weight vector $\emph{\textbf{w}}'_2$ and $\emph{\textbf{w}}_2$. Only if $\alpha=0$, $\theta_{\emph{\textbf{w}}'_2, \emph{\textbf{w}}_2}=0$ and if $\alpha>0$, $\theta_{\emph{\textbf{w}}'_2, \emph{\textbf{w}}_2}>0$. The following discussion is based on $\alpha>0$. If the instance with label 1 is classified correctly, there exists $\|\emph{\textbf{w}}_1\|\|\emph{\textbf{x}}\|\cos{\theta_1}>\|\emph{\textbf{w}}'_2\|\|\emph{\textbf{x}}\|\cos{\theta'_2}$, equivalent to $\cos{\theta_1}>\cos{\theta'_2}$. Because of $\theta_2=\theta'_2+\theta_{\emph{\textbf{w}}'_2, \emph{\textbf{w}}_2}>\theta'_2$, it is satisfied that $\cos{\theta'_2}>\cos{\theta_2}$. Hence, in testing phase, it's satisfied that $\cos{\theta_1}>\cos{\theta'_2}>\cos{\theta_2}$ by a large angular margin. As a result, there are two decision boundaries between any two classes with a large margin.

\textbf{Weights-biased Softmax Loss}. Based on previous discussion, for an instance $\emph{\textbf{x}}$ from class $c$, we evaluate its W-Softmax loss by
\begin{equation}\label{eq:8}
\begin{aligned}
  L&=-\log{\frac{\exp{(\emph{\textbf{w}}_c^T\emph{\textbf{x}})}}{\exp{(\emph{\textbf{w}}_c^T\emph{\textbf{x}})}+\sum_{j\neq{c}}^C\exp{(\emph{\textbf{w}}'^T_j\emph{\textbf{x}})}}} \\
  &=-\log{\frac{\exp{(\emph{\textbf{w}}_c^T\emph{\textbf{x}})}}{\exp{(\emph{\textbf{w}}_c^T\emph{\textbf{x}})}+\sum_{j\neq{c}}^C\exp{(\frac{\alpha\emph{\textbf{w}}_c^T+\emph{\textbf{w}}_j^T}{\|\alpha\emph{\textbf{w}}_c+\emph{\textbf{w}}_j\|}\emph{\textbf{x}})}}},
\end{aligned}
\end{equation}
and it can be simplified as
\begin{equation}\label{eq:9}
  L=-\log{\frac{\exp{(\emph{\textbf{w}}_c^T\emph{\textbf{x}})}}{\sum_{j=1}^C\exp{(\frac{\alpha\emph{\textbf{w}}_c^T+\emph{\textbf{w}}_j^T}{\|\alpha\emph{\textbf{w}}_c+\emph{\textbf{w}}_j\|}\emph{\textbf{x}})}}}.
\end{equation}
Further, for a set of instances $\{\emph{\textbf{x}}_k,k=1,2,\cdots,N\}$, we can evaluate their average W-Softmax loss by
\begin{equation}\label{eq:10}
  L=\frac{1}{N}\sum_k{-\log{\frac{\exp{(\emph{\textbf{w}}_{c_k}^T\emph{\textbf{x}}_k)}}{\sum_{j}\exp{(\frac{\alpha\emph{\textbf{w}}_{c_k}^T+\emph{\textbf{w}}_j^T}{\|\alpha\emph{\textbf{w}}_{c_k}+\emph{\textbf{w}}_j\|}\emph{\textbf{x}}_k)}}}},
\end{equation}
where $c_k$ denotes the label of instance $\emph{\textbf{x}}_k$. Algorithm.\ref{alg:1} summarizes the inference algorithm of CNNs training with W-Softmax loss in one batch input case, where $CELS$ denotes cross entropy loss with softmax.

\begin{algorithm}
\caption{CNNs training with W-Softmax loss in one batch input case.}
\label{alg:1}
\begin{algorithmic}[1]
\REQUIRE Training images $\emph{\textbf{I}}$ with batch size $B$, labels $\emph{\textbf{y}}=[y_1,...,y_B]$;
\STATE Gets extraction network $N_{extr}$ and feature classification network $N_{cls}$ where classifier weight matrix $\emph{\textbf{W}}=[\emph{\textbf{w}}_1,\cdots,\emph{\textbf{w}}_{C}]$;
\STATE Extracting features $\emph{\textbf{X}}=N_{extr}(\emph{\textbf{I}})$ as $\emph{\textbf{X}}=[\emph{\textbf{x}}_1,...,\emph{\textbf{x}}_B]$;
\STATE Initialize total loss $L \gets 0$;
\FOR {$i=1 \to B$}
\STATE Transform classifier weight matrix, gets $\emph{\textbf{W}}_i$ as\\
$\emph{\textbf{W}}_i=[\emph{\textbf{w'}}_1,\cdots,\emph{\textbf{w'}}_{{y_i}-1},\emph{\textbf{w}}_{y_i},\emph{\textbf{w'}}_{{y_i}+1},\cdots,\emph{\textbf{w'}}_{C}]$;
\STATE Calculate instance loss $L_i \gets CELS(\emph{\textbf{W}}_i^T\emph{\textbf{x}}_i,y_i)$;
\STATE Add up instance loss $L \gets L+L_i$;
\ENDFOR
\STATE Update all weights in CNNs, $w \gets w-learning\_rate*\frac{\partial L}{\partial w}$.
\end{algorithmic}
\end{algorithm}

\subsection{Discussion of Hyperparameter \texorpdfstring{$\alpha$}{alpha}}
In the proposed W-Softmax loss, parameter $\alpha$ plays an important role in regulating the decision angular margin. As shown in Fig.\ref{fig:3}, the decision margin will be zero when $\alpha=0$, and in this case, W-Softmax loss becomes the conventional softmax loss. As the value of $\alpha$ increases, the decision margin also increases and the decision boundaries among different classes become more separated. It should be noted that,  for L-Softmax loss and A-Softmax loss, hyper-parameter $m$ is an integer, i.e., $m=2,3,4...$, while hyper-parameter $\alpha$ in our W-Softmax loss is a positive real number ($\alpha\geq{0}$). Although a big value of $\alpha$ can make learned features more discriminative, it also increases the difficulty of training convergence, because it imposes a stronger constraint on the spatial distribution of learned features.

\section{Experiments and Results}
\subsection{Experimental Setting}
\textbf{Datasets}. We carry out the experiments on several standard benchmark datasets, MNIST~\cite{Lecun1998Gradient},CIFAR10~\cite{Krizhevsky2009Learning}, CIFAR100~\cite{Krizhevsky2009Learning} and LFW dataset\cite{Gary2007Lab}. MNIST dataset consists of 60000 binary training images and 10000 binary testing images, and the size of images is $28\times{28}$. Each of datasets CIFAR10 and CIFAR100 consists of 50000 color training images and 10000 color testing images with image size $32\times{32}$. LFW dataset is mainly for face recognition and face verification. In this paper, we focus on face verification part. LFW dataset has 13233 training images covering 5749 people, only 1680 people with two or more images and 6,000 pair images for testing.

\textbf{CNN Setup}. In order to compare expediently with the conventional original softmax loss and other existing losses, we use the CNN architecture presented by~\cite{Liu2016Large} as the backbone. Our experiments are carried with a Quadro P5000 GPU on TensorFlow. In convolution layers, the stride is 1 and PReLU\cite{He2015Delving} is chosen as  the activation function. Momentum optimizer is used in training and the momentum is set to 0.9. We set the initial learning rate as 0.01 for MNIST and CIFAR10/CIFAR100, its exponential decay rate is 0.9 and the decay step is 6000. The weights are initialized by xavier initializer and the weight parameter of weights regularization is 0.0005. Batch normalization is used after PReLU and the dropout is not adopted for the sake of fair comparison.

\textbf{Training Detail}.
To make the network converge quickly in training, we first train the CNNs using the conventional softmax loss, and refine the network using the proposed W-Softmax loss. For LFW dataset, there is an alignment process before training. In our experiment, the faces cropped from all the images are set to 160x160 and the alignment algorithm is adopted from MTCNN\cite{K2016Joint}.

\subsection{Effect of Units' Number \texorpdfstring{$M$}{M} in FC Layer}

\begin{figure}[t]
\centering
\subfigure[]{
\label{fig:mnist}
\begin{minipage}{0.49\linewidth}
\centering
\includegraphics[width=0.98\linewidth]{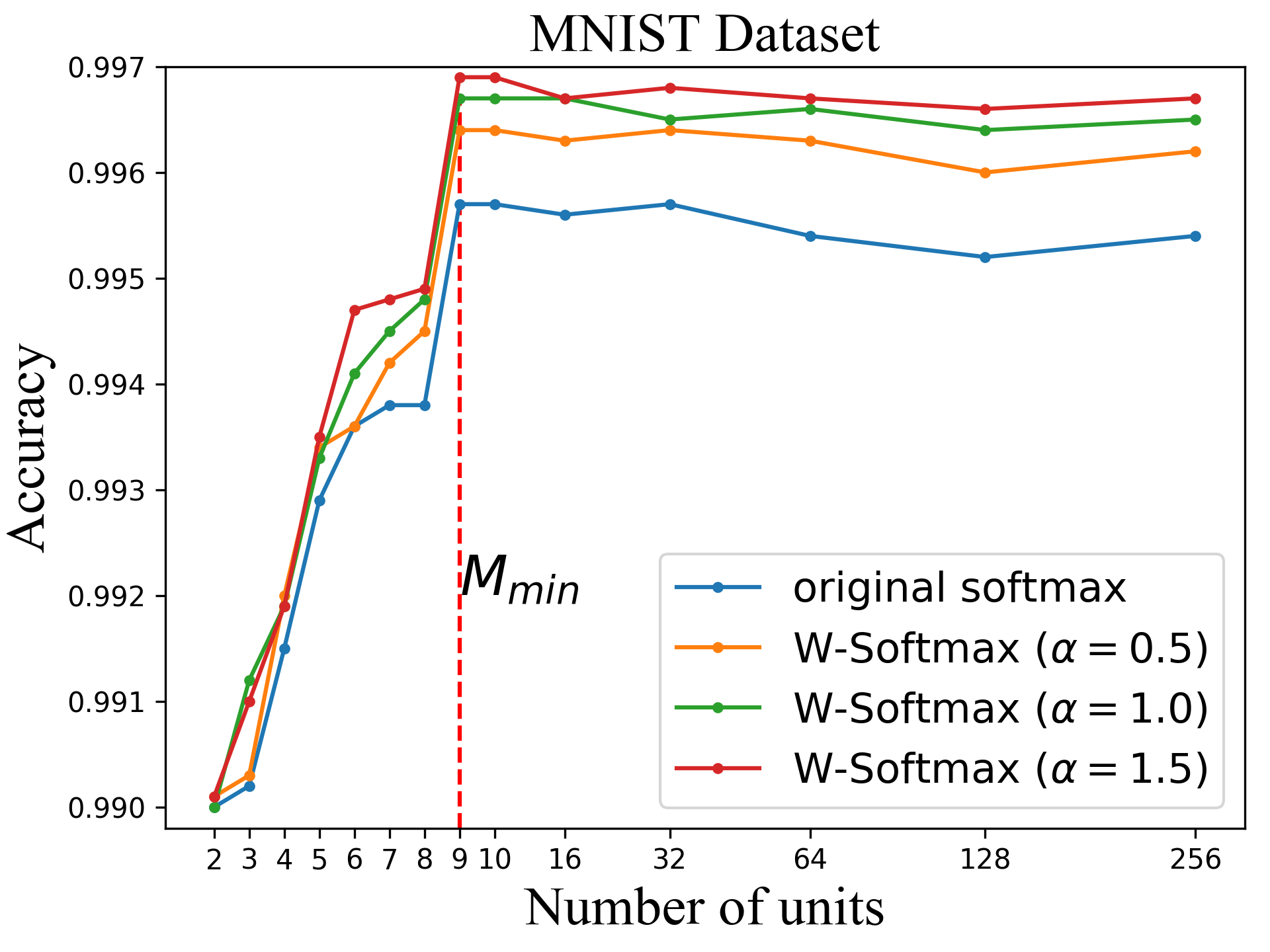}
\end{minipage}%
}%
\subfigure[]{
\label{fig:cifar10}
\begin{minipage}{0.49\linewidth}
\centering
\includegraphics[width=0.98\linewidth]{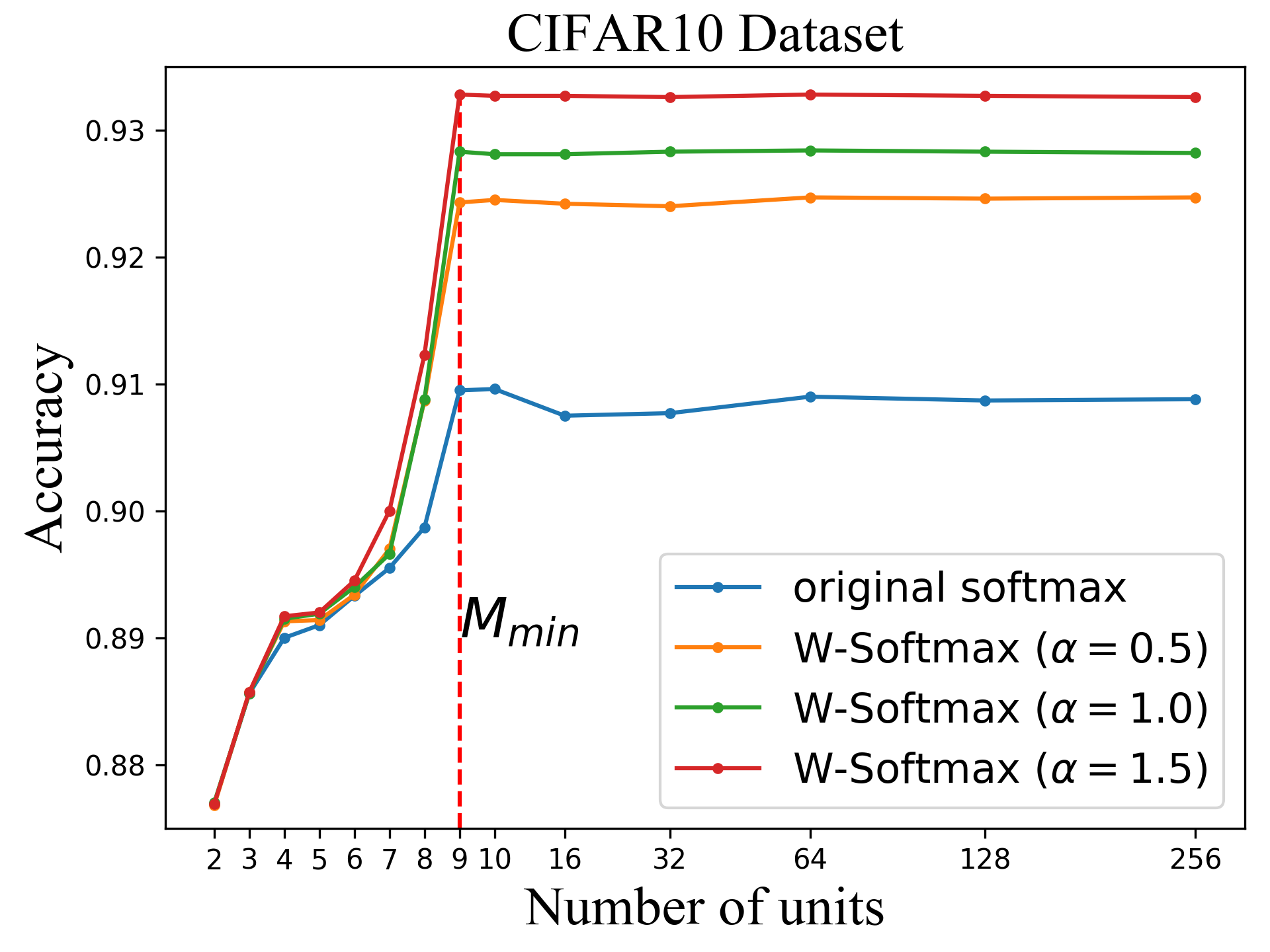}
\end{minipage}%
}%

\subfigure[]{
\label{fig:cifar100}
\begin{minipage}{0.49\linewidth}
\centering
\includegraphics[width=0.98\linewidth]{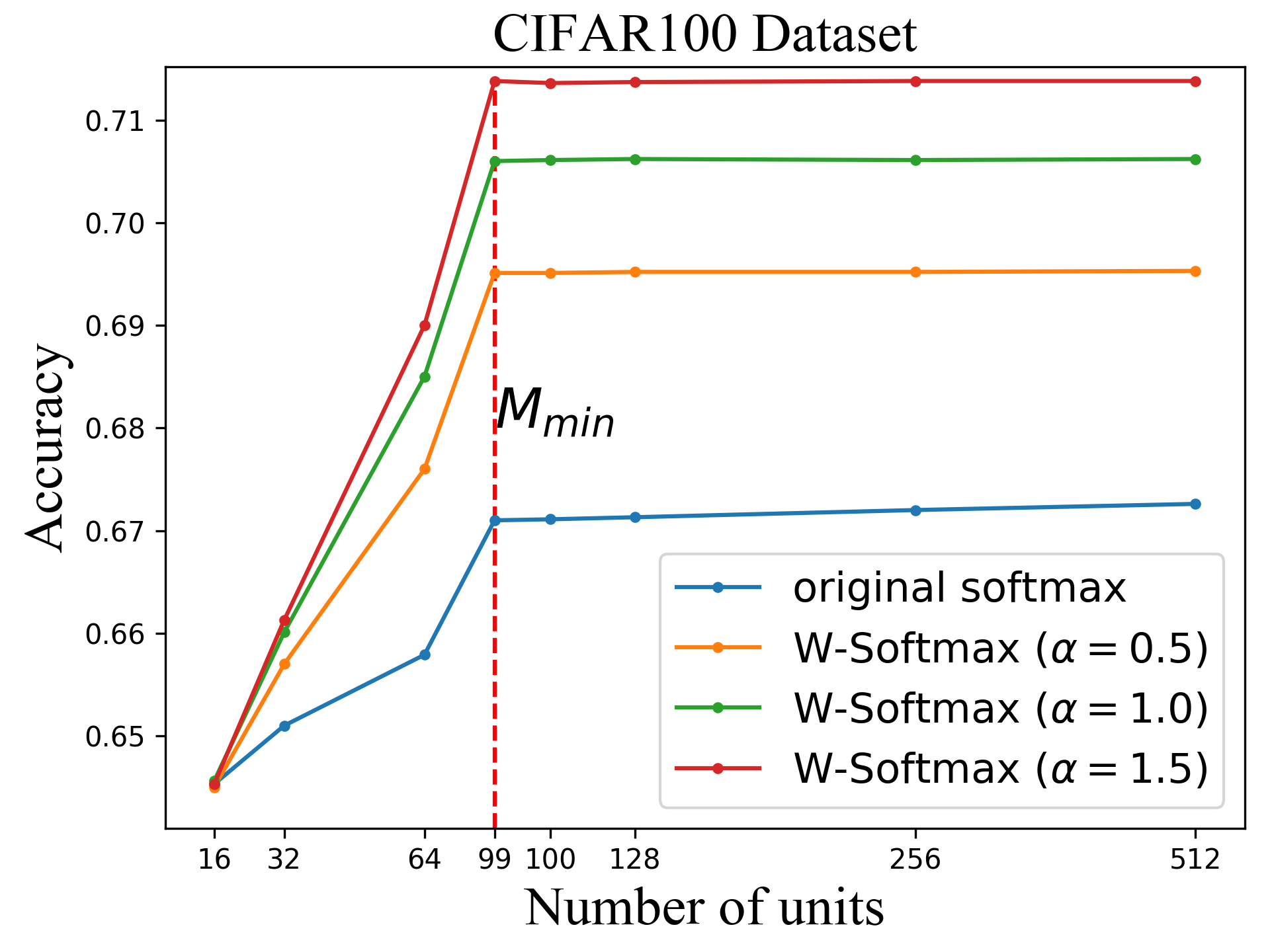}
\end{minipage}
}%
\subfigure[]{
\label{fig:memory}
\begin{minipage}{0.49\linewidth}
\centering
\includegraphics[width=0.98\linewidth]{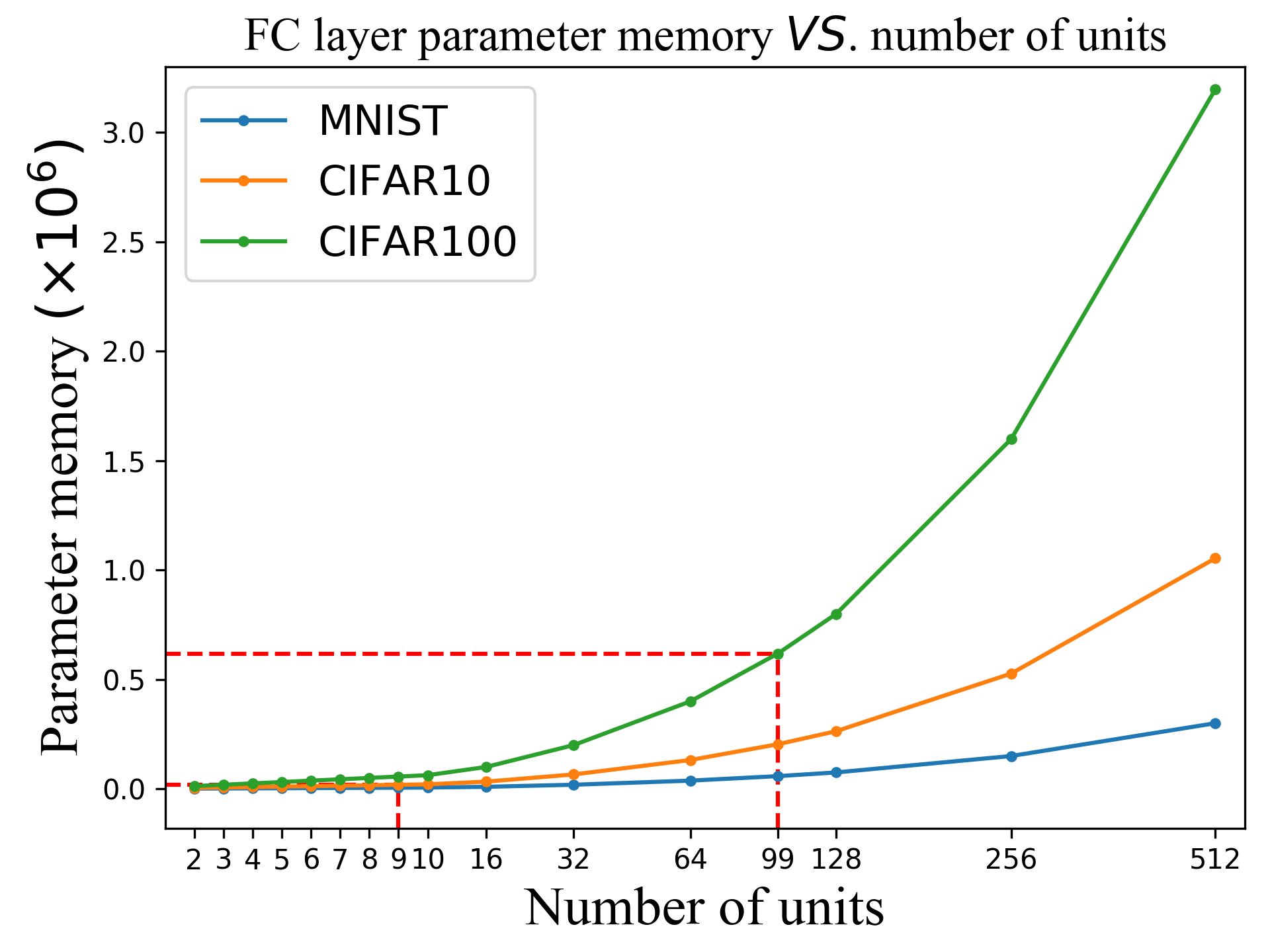}
\end{minipage}
}%
\centering
\caption{Accuracy vs. the number of units $M$ on different datasets with original softmax loss and W-Softmax loss (subfigures (a)-(c)). Subfigure (d) is the parameter memory in FC layer with various number of units. $M_{min}$ is the theoretical minimum proved in previous section.}
\label{fig:number}
\end{figure}

Fig.\ref{fig:mnist}, \ref{fig:cifar10} and \ref{fig:cifar100} illustrate the relation between the number of units in FC layer and classification accuracy on various datasets with different classes. Fig.\ref{fig:memory} is the comparison of the parameter memory in FC layer with different number of units. To visualize the relation detailedly, the scale on the horizontal axis is uneven. The value $M_{min}$ is 9 for MNIST dataset and CIFAR10 dataset and 99 for CIFAR100. Experimental results in Fig.\ref{fig:number} show that when $M<M_{min}$, the accuracies increase with the increase of $M$ and when $M\geq{M_{min}}$, accuracies reach its maximum and fluctuate slightly around it, however, the parameter memory rapidly increases when $M$ gets bigger. Notably, when $M=M_{min}$,
accuracies are close to or even maximum. In MNIST dataset, if $M$ gets to 10x$\sim$30x $M_{min}$, overfitting arises and accuracies with original softmax loss  decline, while the accuracies in CNNs with W-Softmax loss remain almost unchanged, which distinctly eliminate effectively the overfitting.

\subsection{Experiments and Analysis on MNIST}
In the experiments on MNIST, the batch size is set to 50. Table \ref{tab:1} lists the best results of different methods on MNIST. The results in Table \ref{tab:1} and Fig.\ref{fig:mnist} show the proposed W-Softmax loss has better performance than the conventional Softmax loss based on the same network architecture and can achieve the state-of-the-art performance compared with the other methods. The larger $\alpha$ in the W-Softmax loss can bring the higher accuracy to the trained CNN network.
\begin{table}[ht]
  \begin{center}
  \caption{Test accuracy(\%) of different methods on MNIST, where * denotes our proposed method.}\label{tab:1}
  \begin{tabular}{|c|c|}
  \hline
  Method & test accuracy(\%) \\\hline
  \hline
  CNN\cite{K2009What} & 99.47 \\
  DropConnect\cite{Wan2013Regularization} & 99.43 \\
  FitNet\cite{Romero2015Fit} & 99.49 \\
  NiN\cite{Min2014NIN} & 99.53 \\
  Maxout\cite{Goodfellow2013Maxout} & 99.55 \\
  DSN\cite{Lee2015Deep} & 99.61 \\
  R-CNN\cite{Ming2015Rec} & \textbf{99.69} \\
  GenPool\cite{Lee2015General} & \textbf{99.69} \\
  \hline
  Hinge Loss & 99.53 \\
  original softmax & 99.58 \\
  L-Softmax\cite{Liu2016Large} & \textbf{99.69} \\\hline
  W-Softmax($\alpha$=0.5)* & 99.64 \\
  W-Softmax($\alpha$=1)* & 99.67 \\
  W-Softmax($\alpha$=1.5)* & \textbf{99.69} \\\hline
  \end{tabular}
  \end{center}
\end{table}

\begin{figure}[t]
  \centering
  \includegraphics[width=1.0\linewidth]{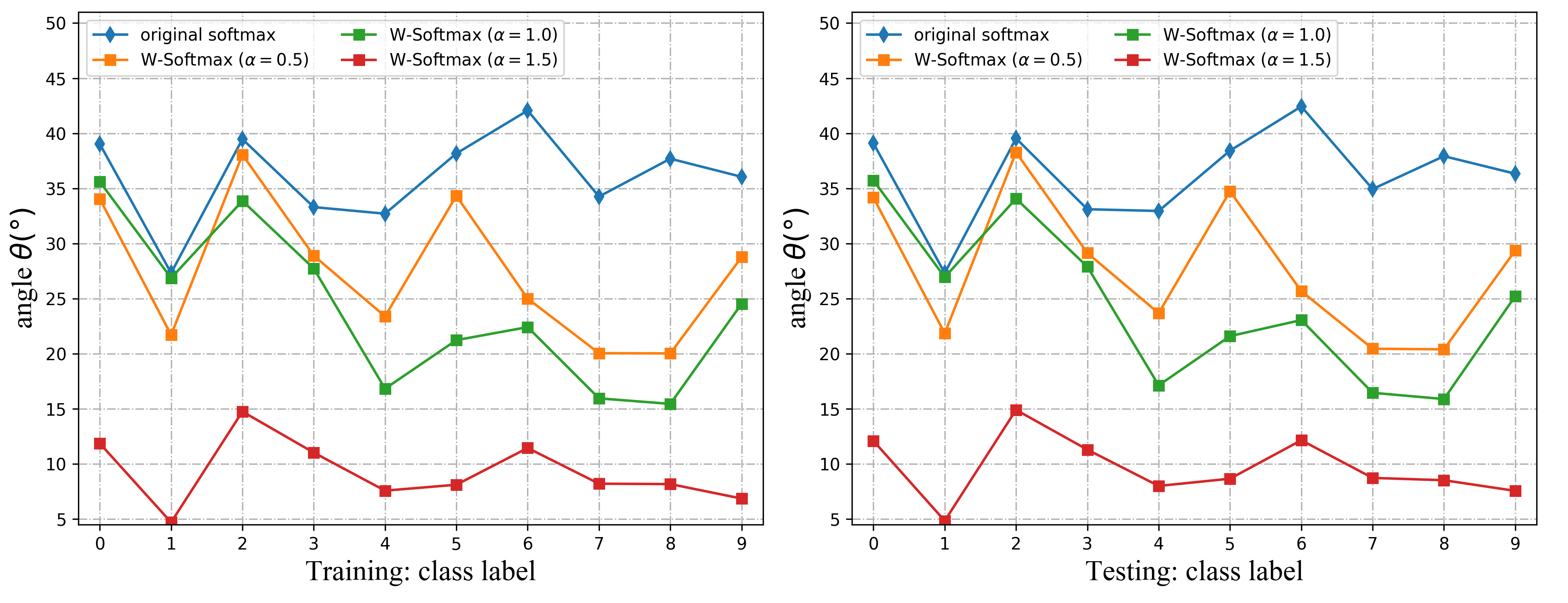}\\
  \caption{Learned features comparison between original softmax loss and W-Softmax loss on MNIST dataset. The value in vertical axis is the mean angle $\overline{\theta}_i$ between $\textbf{\emph{w}}_i$ and all $\textbf{\emph{x}}_i$ for class label $i$. The left figure is the result on training dataset and the right one is on testing dataset.}\label{fig:angle}
\end{figure}

\textbf{Learned Features Comparison}. We calculate the angles between learned features and the weight vectors of classifiers corresponding to their true categories,  and then get the mean of these angles, i.e.,
\begin{equation}\label{eq:theta}
  \overline{\theta}_i=\frac{1}{N_i}\sum_{j=1}^{N_i}{\arccos{\frac{\textbf{\emph{w}}_i^T\textbf{\emph{x}}_i^{(j)}}{\|\textbf{\emph{w}}_i\|\|\textbf{\emph{x}}_i^{(j)}\|}}},\forall{i}=1,2,\cdots,C
\end{equation}
where $C$ is the number of classes in dataset, $N_i$ is the number of instances with label $i$, $\textbf{\emph{x}}_i^{(j)}$ denotes the $j$-th instance of class $i$, $\textbf{\emph{w}}_i$ is the weight vector of classifier responsible for class $i$ and $\overline{\theta}_i$ denotes the mean of angles between $\textbf{\emph{w}}_i$ and all $\textbf{\emph{x}}_i^{(j)}$. The statistical results are shown in Fig.\ref{fig:angle}, where lower mean angle corresponds to the compactness of intra-class. We can see that the mean angle of each class with original softmax loss is larger than those in W-Softmax loss with different $\alpha$, which denotes conventional softmax loss can not encourage the intra-class compactness. It's conspicuous that the mean angle gets smaller when $\alpha$ increases. The mean angle of training dataset is slightly less than the one of testing dataset for each class, and larger $\alpha$ in the W-Softmax loss can encourage the intra-class compactness and inter-class separability.

\subsection{Experiments on CIFAR10 and CIFAR100}
Both CIFAR10 and CIFAR100 have 50000 training instances and 10000 testing instances. But, CIFAR10 has 5000 training instances for each class, and CIFAR100 only has 500. In the training, batch size is 256 for CIFAR10 and CIFAR100. The experimental results in Table \ref{tab:3} show that the W-Softmax loss achieves 2\%-3\% improvement on CIFAR10 and improves more than 4\% accuracy on CIFAR100 compared with the conventional softmax loss.
\begin{table}[ht]
  \begin{center}
  \caption{Test accuracy(\%) of different methods on CIFAR10 and CIFAR100, where N/A means the lack of comparative results.}\label{tab:3}
  \begin{tabular}{|c|c|c|}
  \hline
  Method & CIFAR10(\%) & CIFAR100(\%) \\\hline
  \hline
  DropConnect\cite{Wan2013Regularization} & 90.59 & N/A \\
  FitNet\cite{Romero2015Fit} & N/A & 64.96 \\
  NiN\cite{Min2014NIN} & 89.53 & 64.32 \\
  Maxout\cite{Goodfellow2013Maxout} & 88.32 & 61.43 \\
  DSN\cite{Lee2015Deep} & 90.31 & 65.43 \\
  All-CNN\cite{To2015Striving} & 90.92 & 66.29 \\
  R-CNN\cite{Ming2015Rec} & 91.31 & 68.25 \\
  GenPool\cite{Lee2015General} & 92.38 & 67.63 \\
  \hline
  Hinge Loss & 90.09 & 67.10 \\
  original softmax & 90.95 & 67.26 \\
  L-Softmax\cite{Liu2016Large} & \textbf{92.42} & \textbf{70.47} \\\hline
  W-Softmax($\alpha$=0.5)* & 92.47 & 69.53 \\
  W-Softmax($\alpha$=1)* & 92.84 & 70.62\\
  W-Softmax($\alpha$=1.5)* & \textbf{93.28} & \textbf{71.38} \\\hline
  \end{tabular}
  \end{center}
\end{table}

\subsection{Experiments on LFW Dataset}
The faces in LFW dataset are detected and aligned by MTCNN\cite{K2016Joint} and then cropped to 160x160. Before training and testing, each face image is normalized to $[-1,1]$ by subtracting
127.5 and then dividing by 128. The feature extraction network is trained on a small training dataset that is the publicly available CASIA-WebFace\cite{Dong2014Le} dataset containing 0.49M face images from 10,575 subjects. When training feature extraction network, batch size is set to 128 and the learning rate is initially 0.1 and divided by 10 for every 10k iterations, and training is stopped at 30k iterations. The cosine distance of features is adopted as the similarity score. The result is shown in Tabel \ref{tab:LFW}. Compared with the original softmax loss, the accuracy of W-SoftMax loss is greatly improved, which proves that encouraging intra-class compactness and inter-class separability is more conducive to  improving the accuracy of face verification.
\begin{table}[ht]
  \begin{center}
  \caption{Face verification (\%) on the LFW dataset, where * denotes the outside data is private (not publicly available).}\label{tab:LFW}
  \begin{tabular}{|c|c|c|}
  \hline
  Method & Outside Data & Accuracy(\%) \\\hline
  \hline
  FaceNet\cite{F2015FaceNet} & 200M* & \textbf{99.65} \\
  Deep FR\cite{O2015Deep} & 2.6M & 98.95 \\
  DeepID2+\cite{Y2015Deeply} & 300K* & 98.97 \\
  L-Softmax\cite{Liu2016Large} & WebFace & \textbf{98.71} \\
  \hline
  original softmax & WebFace & 96.53 \\
  \hline
  W-Softmax($\alpha$=0.5) & WebFace & 97.98 \\
  W-Softmax($\alpha$=1) & WebFace & 98.86\\
  W-Softmax($\alpha$=1.5) & WebFace & \textbf{98.91} \\\hline
  \end{tabular}
  \end{center}
\end{table}

\subsection{Experiments with Multi-class}
To explore the effect of W-Softmax when the number of classes increases, we design the experiments on MNIST and CIFAR10. Concretely, we randomly select $k$ classes from 10 classes. Since the accuracy on MNIST almost reaches 100\% when $k=2$, we choose $k$ from 5 to 10, and specially select the first $k$ classes from 10 classes and set $\alpha=1$ in our experiments. The experimental results in Table \ref{tab:2} show that (1) the classification accuracy decreases as $k$ increases; (2) when $k$ is small, the advantage of W-Softmax loss over conventional Softmax is not obvious, and when $k$ is big enough for a specific dataset, a big gain can be obtained.
\begin{table}[ht]
  \begin{center}
  \caption{Effect of W-Softmax to multi-class number on MNIST and CIFAR10. The 'softmax' in table denotes original softmax loss.}\label{tab:2}
  \begin{tabular}{|c|cc|cc|}
  \hline
  \multirow{2}*{\tabincell{c}{\small{class}\\\small{number}}} & \multicolumn{2}{|c|}{MNIST} & \multicolumn{2}{|c|}{CIFAR10} \\\cline{2-5}
  & \small{softmax} & \small W-Softmax & \small{softmax} & \small W-Softmax \\\hline
  $k$=5 & 99.92 & 99.93 & 94.35 & 95.23 \\
  $k$=6 & 99.83 & 99.88 & 91.55 & 93.08 \\
  $k$=7 & 99.67 & 99.78 & 91.26 & 92.65 \\
  $k$=8 & 99.67 & 99.76 & 91.27 & 92.73 \\
  $k$=9 & 99.63 & 99.73 & 91.24 & 92.72 \\
  $k$=10 & 99.58 & 99.69 & 90.95 & 92.44 \\\hline
  \end{tabular}
  \end{center}
\end{table}

\section{Conclusion}
In this paper, we theoretically determine the minimum number of nodes of classifier weight and verify this by experiments, which reduces the CNNs' parameter and training time. We present a new weights-biased Softmax(W-Softmax) loss, which is useful to build high-performance CNNs by learning highly discriminative features. By applying it, the decision margin can be flexibly adjusted by parameter $\alpha$. The preliminary experiments show W-Softmax loss can achieve obvious improvement over conventional Softmax loss and obtain comparable or better classification accuracy in CNNs training compared with state-of-the-art loss functions.

\section*{References}

\bibliography{mybibfile}

\end{document}